\UseRawInputEncoding
\documentclass[conference]{IEEEtran}
\usepackage{cite}
\usepackage{amsmath,amssymb,amsfonts}
\usepackage{algorithmic}
\usepackage{graphicx}
\usepackage{textcomp}
\usepackage{xcolor}
\def\BibTeX{{\rm B\kern-.05em{\sc i\kern-.025em b}\kern-.08em
    T\kern-.1667em\lower.7ex\hbox{E}\kern-.125emX}}
\usepackage[square,numbers]{natbib}
\usepackage[hyphens]{url}
\begin{document}

\title{Proposing a Framework for Machine Learning Adoption on Legacy Systems}

\author{\IEEEauthorblockN{1\textsuperscript{st} Ashiqur Rahman}
\IEEEauthorblockA{\textit{Department of Computer Science} \\
\textit{Northern Illinois University}\\
DeKalb, IL, USA \\
https://orcid.org/0000-0003-3290-2474}
\and
\IEEEauthorblockN{2\textsuperscript{nd} Hamed Alhoori}
\IEEEauthorblockA{\textit{Department of Computer Science} \\
\textit{Northern Illinois University}\\
DeKalb, IL, USA \\
https://orcid.org/0000-0002-4733-6586}
}

\maketitle

\begin{abstract}
The integration of machine learning (ML) is critical for industrial competitiveness, yet its adoption is frequently stalled by the prohibitive costs and operational disruptions of upgrading legacy systems. The financial and logistical overhead required to support the full ML lifecycle presents a formidable barrier to widespread implementation, particularly for small and medium-sized enterprises. This paper introduces a pragmatic, API-based framework designed to overcome these challenges by strategically decoupling the ML model lifecycle from the production environment. Our solution delivers the analytical power of ML to domain experts through a lightweight, browser-based interface, eliminating the need for local hardware upgrades and ensuring model maintenance can occur with zero production downtime. This human-in-the-loop approach empowers experts with interactive control over model parameters, fostering trust and facilitating seamless integration into existing workflows. By mitigating the primary financial and operational risks, this framework offers a scalable and accessible pathway to enhance production quality and safety, thereby strengthening the competitive advantage of the manufacturing sector.
\end{abstract}


\section{Introduction}\label{introduction}
In the competitive landscape of modern manufacturing, shaped by the fourth industrial revolution, the pursuit of efficiency is paramount \citep{Radziwon2014-if, Muller2018-kw, Rao2022-xv, Soni2023-sy}. This creates a central tension between the need to increase production speed and upholding the rigorous safety and quality standards that define the industry. Machine Learning (ML) and Artificial Intelligence (AI) present a transformative solution, offering powerful tools to optimize production and enhance quality control. However, a significant gap exists between the promise of these technologies and their practical implementation, largely due to prohibitive integration barriers. This challenge is acutely felt in safety-critical processes like Non-Destructive Inspection (NDI), the evaluation of components without causing damage, where the stakes for accuracy and reliability are highest \citep{Honarvar2020-tw, Ould_Naffa2002-ab, Dwivedi2018-ts, Memmolo2015-er, Brosnan2004-yd}.

As a cornerstone of quality assurance, NDI has traditionally relied on the meticulous work of human experts, who manually inspect signals from ultrasound, X-ray, or other scans to identify potential flaws \citep{Yang2013-mh, Zhang2020-kp}. While essential, this process is inherently repetitive and cognitively demanding, making it susceptible to errors born from fatigue. These errors can have severe consequences for product quality and safety. It is precisely this challenge that ML models are poised to address. By training algorithms to detect anomalies in scanned signals, ML can serve as a powerful assistive tool, augmenting the capabilities of human inspectors by rapidly flagging potential defects for review \citep{Gardner2020-bn, Taheri2022-ix}. The potential benefits are clear and substantial, including reduced inspection times, lower operational costs, and a significant mitigation of human error.

Despite this compelling value proposition, the widespread adoption of ML in manufacturing is often hindered by a formidable set of practical and financial hurdles. The computational demands of training and executing complex ML models far exceed the capabilities of the legacy systems prevalent across the industry \citep{Taheri2022-ix}. Consequently, organizations are faced with a daunting trade-off between either undertaking a prohibitively expensive and disruptive overhaul of their core infrastructure or forgoing the competitive advantages of ML. The primary barriers to adoption are well-documented and include high upfront capital expenditure on hardware, the operational risk of destabilizing critical production workflows, the necessity for extensive personnel retraining, and the unacceptable cost of production downtime during installation and maintenance \citep{Hayretci2021-xy}. For small and medium-sized enterprises (SMEs), these obstacles are often insurmountable, widening the competitive gap. Therefore, a viable integration strategy is necessary to specifically dismantle these barriers, offering a pathway to ML adoption that is cost-effective, non-disruptive, and user-centric.

To address these challenges, this paper introduces a framework built upon a proven, API-centric architecture, an industry-standard pattern for modernizing legacy systems with minimal risk. The core of our approach is the strategic decoupling of the ML model lifecycle from the on-site production environment. This separation is the key mechanism that dismantles the barriers to adoption. By hosting models on a separate, modern infrastructure, all model maintenance, retraining, and updates can occur with zero production downtime and without requiring any costly upgrades to existing legacy hardware. The benefits are delivered to domain experts through an intuitive, lightweight, browser-based interface. This tool is designed not to replace human expertise, but to augment it, promoting a collaborative relationship between the inspector and the AI. It empowers users with direct control over model parameters, such as sensitivity and confidence thresholds, allowing them to interact with the model's predictions in real time. This interactivity is crucial in transforming the ML model from an opaque ``black box" into a transparent and trustworthy assistive tool, fostering the user confidence essential for adoption.

While this paper uses the NDI process as a primary motivating example, the proposed framework is fundamentally generalizable. It provides a pragmatic and scalable blueprint for integrating advanced ML capabilities into any industrial sector where legacy systems and operational continuity are critical concerns. By offering a low-cost, non-disruptive, and human-centered pathway to adoption, this work aims to bridge the deployment gap and unlock the full potential of machine learning for the broader manufacturing industry.

\section{Related Works}\label{related-works}
\subsection{Use and Challenges of AI and ML in Production}\label{ai-ml-in-production}
For an AI system to be successful in a production environment, it must be sustainable. This requires it to remain operational, continuously learn from new data, and improve itself while staying true to its core purpose \citep{Tamburri2020-ee}. A critical component of this sustainability is the process of continuous re-training, which keeps the model updated in a changing world. This re-training cycle is resource-intensive, demanding robust data processing pipelines, significant computational power, and ongoing evaluation to ensure system accuracy \citep{Wazir2023-uv}.

The process of updating a production-level ML model introduces a cascade of challenges for data and ML engineers. These include managing new computational requirements, maintaining system coherence, and upholding quality assurance standards \citep{Makinen2021-jb}. Furthermore, practical issues such as accessing new data and the high costs associated with computational resources for re-training and evaluation can be significant concerns for industry executives \citep{Taheri2022-ix}. The long-term value of an AI system is therefore dependent on a continuous financial commitment. The limited interpretability of complex deep learning models also presents a significant hurdle, particularly in safety-critical environments where understanding the decision-making process of the model is paramount \citep{Lin2023-fd}. These major concerns surrounding data management, re-training, and security necessitate further research to streamline the deployment of ML models in production \citep{Paleyes2022-eg}.

A small body of research has explored the use of APIs as a potential solution to some of these challenges. The DEEPaaS API, for example, proposes a Python-based endpoint for developers to expose the functionalities of their models to end users \citep{Lopez_Garcia2019-ds}. Similarly, \citet{Hitimana2024-ql} presented a web and mobile application that uses a deep learning model in the backend to detect diseases in coffee crops, showcasing a method for integrating a consumer-facing application with a remote ML model. However, these approaches often lack a robust infrastructure for model maintenance, which could lead to service downtime during model upgrades. Additionally, existing API-based solutions typically do not provide an all-around solution that includes the ability for end-users to control and interact with model parameters.

\subsection{ML in the Industry}
The growth of ML models across manufacturing, security, and numerous other industries is undeniable. The application of ML to the NDI process is a particularly active area of research, where experts inspect non-destructive scans to identify regions of interest. Numerous researchers have proposed a variety of ML techniques to improve and automate the NDI process. Comprehensive review papers by \citet{Gholizadeh2016-ww}, \citet{Chauveau2018-qo}, and \citet{Dwivedi2018-ts} have surveyed the advancements in non-destructive testing techniques available to the industry. Many of these techniques are deployed in safety-critical sectors such as aerospace and healthcare \citep{Dwivedi2018-ts, An2023-jm}. In a related context, \citet{Brosnan2004-yd} reviewed technologies for the food industry and highlighted the importance of computer vision systems for inspection processes.

While ML has proven its value in handling large volumes of data and offering high accuracy \citep{Gardner2020-bn, Taheri2022-ix}, its broader adoption is hindered by persistent challenges. These obstacles include high computational costs, a lack of available training data, the difficulty of appropriate hyperparameter tuning, and the persistent issue of limited model interpretability \citep{Taheri2022-ix, Lin2023-fd, Sun2023-yp}. These challenges represent significant barriers to the widespread implementation of ML in industrial settings. Our proposed framework seeks to directly tackle several of these issues to lower the barrier of entry and promote the wider use of ML.

\subsection{Challenges for Legacy Systems}
In the era of Industry 4.0, capabilities such as large-scale data processing, connectivity, automation, AI, and agility are key factors for success \citep{Muller2018-kw, Rao2022-xv, Soni2023-sy}. However, the high cost and complexity of implementation remain major barriers preventing small and medium enterprises (SMEs) from adopting Industry 4.0 technologies \citep{Radziwon2014-if}. The literature suggests that when adopting ML, companies must evaluate factors beyond model accuracy, including implementation costs, generalizability, and the necessary workforce competence \citep{Rana2014-dn}. Broader discussions on the challenges of AI adoption highlight technological barriers, concerns about job displacement, the need for significant investment, and even political willingness as key factors \citep{Rao2022-xv}. The gap in technological skills, high computational costs, and concerns about data security are also frequently cited as primary challenges for AI implementation in industry \citep{Soni2023-sy}.

The literature confirms that even when organizations commit to modernization, the process of upgrading legacy systems is fraught with difficulty. The promise of increased efficiency and security often confronts a complex web of both technical and organizational obstacles. From a technical standpoint, modernization projects frequently encounter challenges that include system compatibility issues, resource shortages, necessary process changes, and insufficient training \citep{Mendonca2023-vc}. These are often accompanied by deeper-rooted problems such as intricate system dependencies, difficult data migrations, substantial technical debt, and the need to maintain regulatory compliance \citep{Chinamanagonda2024-sw}. Beyond these technical hurdles, a recurring and critical theme in the literature is the challenge of organizational and user resistance. Significant obstacles arise from a natural fear of change, concerns about production downtime, and the direct impact on end-users, with user acceptance being identified as a crucial and often overlooked factor for success \citep{Ponnusamy2023-lu}. When the modernization specifically involves machine learning, these challenges are further compounded by the high cost of data acquisition and the essential need for interpretable and accountable ML-driven decisions \citep{Khan2022-pv}.

The framework proposed in this paper is designed as a direct response to this well-documented landscape of challenges. It is engineered to facilitate adoption by mitigating costs, increasing user control to improve accountability, and removing the necessity for the direct technological modernization of existing legacy systems.

\section{Method}
\begin{figure}
    \centering
    \includegraphics[width=1\linewidth]{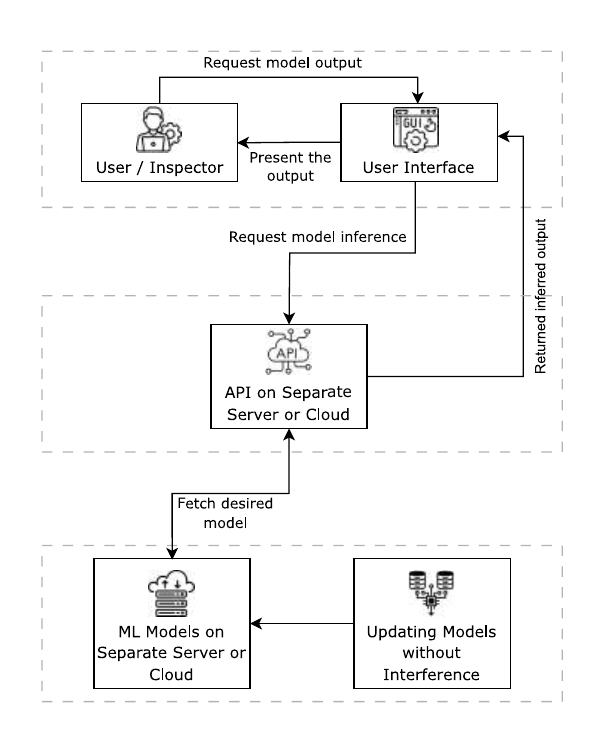}
    \caption{The components of the proposed method. Each component is surrounded by dotted lines. The one at the top represents the user interface on a legacy system, then the one in the middle represents the API, and the one at the bottom represents the ML model maintenance.}
    \label{fig:components}
\end{figure}

The framework we propose consists of three core components designed to integrate advanced ML capabilities into existing industrial workflows. These key parts are a centralized repository for ML models, a lightweight and intuitive user interface (UI), and a robust Application Programming Interface (API) that functions as the communication bridge between the UI and the models. The interaction and workflow between these components are illustrated in Figure \ref{fig:components}.

\subsection{ML Model Hosting and Management}
A key principle of our architectural approach is the complete decoupling of the ML model lifecycle from the operational manufacturing pipeline. All stages of model management, including training, validation, and deployment, are conducted within an isolated computational environment that is physically and logically separate from the on-site legacy systems. This environment, which can be a dedicated on-premises server or a third-party cloud service, can be equipped with the modern hardware and software stacks necessary for computationally intensive ML tasks.

This strategic separation yields several critical advantages that directly address the primary barriers to ML adoption. It ensures that the entire model lifecycle can be managed with zero downtime or disruption to the production floor. This isolation is particularly beneficial during model re-training with new data. The training and validation of an updated model version can occur in a parallel workflow, completely independent of the currently deployed model that is serving live production requests. The existing, stable model remains fully operational and unaffected while the next iteration is developed and tested. Once the new model is approved, it can be deployed, and the API gateway can seamlessly redirect traffic to the updated version with no interruption to the end-user. This agile approach to model management eliminates the risk and complexity associated with on-device updates. Furthermore, this architecture completely obviates the need for any hardware or software upgrades to the legacy client systems, preserving the stability of established workflows.

Leveraging a cloud environment for model hosting offers further strategic benefits, particularly from a financial perspective. It allows an organization to convert what would be a significant capital expenditure on specialized on-site hardware into a more manageable operational expenditure. This pay-for-use model provides greater financial flexibility and scalability, allowing computational resources to be scaled up or down based on demand without the burden of purchasing and maintaining powerful, and often underutilized, systems in-house.

\subsection{User-Centric Interface Design}
The user interface is fundamentally designed to empower domain experts, functioning as an intelligent assistive tool that augments rather than replaces their critical judgment. Our prototype, which runs in any standard web browser, is built on a philosophy of human-in-the-loop collaboration. It provides users with the autonomy to select the appropriate dataset and choose the most suitable ML model for a specific inspection task. This step ensures that the invaluable experience and contextual knowledge of the expert guide the analytical process from the start. The design intentionally places the human operator in a position of authority, ensuring the technology serves their workflow instead of dictating it.

\begin{figure*}[tb]
    \centering    \frame{\includegraphics[width=0.9\linewidth]{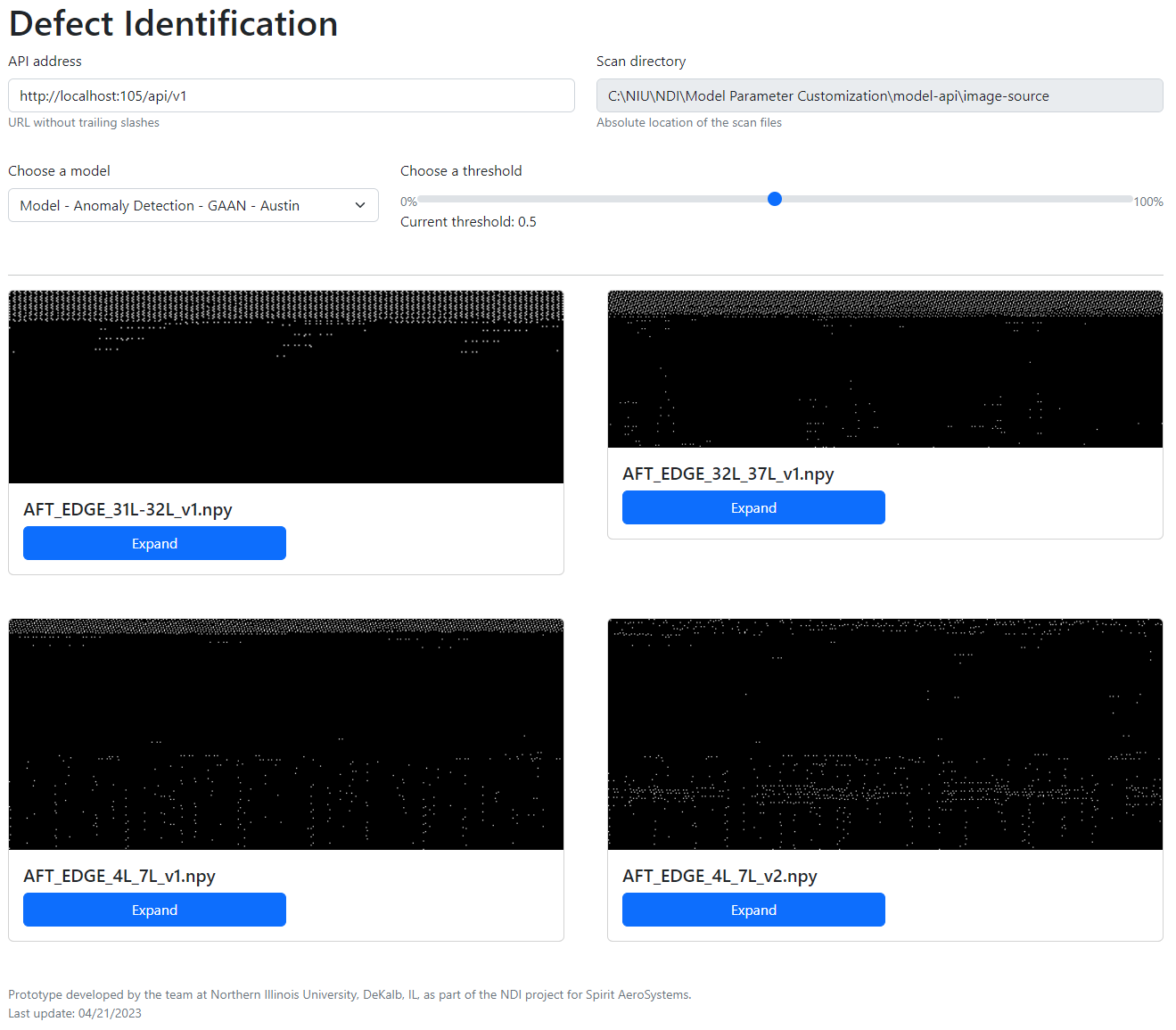}}
    \caption{Prototype user interface for the framework. The white dots on the black scan images represent regions of interest identified by the model.}
    \label{fig:va-tool}
\end{figure*}

A key feature of the interface is the ability for users to interactively adjust model parameters, such as sensitivity or confidence thresholds, prior to running an inference. This interactivity is crucial for demystifying the functionality of the model and building the user's trust in its outputs. By allowing experts to probe the model and observe how its predictions change in real time, the interface transforms the algorithm from an opaque ``black box" into a transparent and understandable tool. Upon completion of the inference task, the interface presents the results in a clear, visual format. This allows the user to efficiently inspect the machine-generated output and draw their own informed conclusions, reinforcing the role of the expert as the final decision-maker and facilitating a smoother integration into established quality assurance protocols.

This user-centric approach is supported by a design that prioritizes accessibility and minimizes deployment friction. The minimal system requirements, which include only a standard web browser and a network connection to the API endpoint, ensure the tool can be adopted with negligible cost overhead and disruption. By leveraging a universally familiar browser-based paradigm, the framework obviates the need for complex software installations on client machines and significantly reduces the learning curve for personnel. This ensures that the focus remains on the inspection task itself, not on learning a new and complicated piece of software, thereby promoting wider and more rapid adoption.

\subsection{The API Gateway}
The API gateway serves as the keystone of the framework, functioning as the central communication hub that connects the user-facing interface with the back-end model repository. It orchestrates the entire inference workflow from initiation to completion. The process begins when the API receives a request from the user, which bundles the input data, the identifier for the expert-selected model, and any user-adjusted parameters. The gateway is then responsible for securely routing this request to the appropriate model hosted in the remote computational environment, executing the inference task, and managing the response.

Once the model has processed the data, the API receives the raw output, formats the results into a standardized structure, and transmits them back to the user interface for clear visual presentation. This role as a single point of entry is also critical for ensuring robust security and manageability. All interactions are funneled through the gateway, which allows for centralized control over access, authentication, and request logging, thereby safeguarding valuable intellectual property and providing a clear audit trail. To minimize data transfer latency in our prototype and ensure a responsive user experience, all data is exchanged using the efficient NumPy array format, which is well-suited for the numerical data common in ML tasks. The architecture remains flexible enough to accommodate other data structures as needed.

A functional prototype of the proposed system, which encompasses all three core components, has been made publicly available on GitHub\footnote{https://github.com/DATALab-NIU/legacy-ml-integration}. This is intended to encourage full reproducibility of our work and to facilitate further research and development by the broader community.

\section{Discussion}
The integration of ML models into industrial workflows presents a critical challenge that often impedes the translation of academic advancements into measurable increases in real-world productivity. While human expertise remains the cornerstone of any industrial processes, such as product development and quality assurance, personnel are often subject to significant strain. This can lead to fatigue and an increased propensity for error. ML models are suitable to serve as powerful assistive tools, augmenting the capabilities of human experts to enhance both productivity and quality control. However, despite a wealth of research into specialized industrial ML applications, their deployment within real-world production environments remains an obstacle.

The core of this challenge lies in the operational and technical friction between cutting-edge ML technologies and established legacy systems. Effective ML models necessitate a continuous lifecycle of updates, retraining, and maintenance to adapt to evolving data distributions and operational requirements. This lifecycle imposes significant computational and logistical overhead, often demanding hardware and software infrastructure that legacy systems cannot support. Consequently, organizations face a difficult choice between either undertaking costly and disruptive overhauls of their existing production pipelines or forgoing the benefits of ML altogether.

To overcome this deployment challenge, we propose a framework that separates the ML model's operations from the end-user's computer by using a central, API-based service. This approach hides the model's complex calculations and delivers its benefits to the user as a simple, accessible tool. Instead of dealing with a complicated local software installation, the human expert uses a lightweight, browser-based interface that connects to a remote API. This simple setup requires only a standard web browser and a network connection to the API, which removes the need to modify existing legacy systems and greatly reduces the requirement for personnel retraining. While a formal user study to empirically validate this specific implementation was outside the scope of this work, the framework's practical feasibility and high potential for user acceptance can be strongly inferred from a convergence of evidence across its architectural design, its economic value proposition, and its alignment with established human-computer interaction (HCI) principles.

This approach completely rethinks model deployment. The large, computationally demanding ML models are hosted on specialized, modern infrastructure that is separate from the day-to-day operational systems. This infrastructure can be located on-premises or, more significantly, in the cloud. Opting for a cloud-based solution introduces substantial cost benefits, as it allows organizations to pay only for the computational resources they use, eliminating the need to purchase and maintain expensive, powerful hardware in-house. This centralized system then serves as the single point for all model-related operations, from running predictions to performing necessary updates and maintenance. The proposed API-centric architecture mirrors validated, best-practice patterns for legacy system modernization that have been successfully adopted across multiple mission-critical sectors. This approach of using an API as an abstraction layer is the de facto standard in the highly regulated financial services and healthcare industries, where it is trusted to securely connect monolithic core systems to modern applications without compromising stability or compliance \citep{Faruk2022-rf, Koneni2024-hl, Koritala2025-uh}. This extensive cross-industry precedent provides a strong validation of the architecture's technical robustness and reliability.

Additionally, the proposed API-based method offers a range of benefits that address the main barriers to industrial ML adoption. A key benefit is that this framework empowers organizations to stay technologically current without costly technological investment. By offloading the computational burden of ML inference to a remote server, our approach avoids the need for costly upgrades to legacy systems. This immediately lowers the barrier to entry for industries reliant on established, validated, and often aging infrastructure, allowing them to leverage state-of-the-art AI without disrupting their stable production environments. The feasibility of the framework is underwritten by a powerful economic imperative as well. The manufacturing sector has already demonstrated the transformative value of ML, with industry leaders reporting quantifiable impacts such as a $40\%$ reduction in manufacturing flaws by BMW and a $40\%$ decrease in unplanned downtime by General Electric \citep{Asamaka-Industries-Ltd2025-qt}. This proven, multi-million-dollar value creates a significant economic motivation for adopting low-friction deployment models that can overcome well-documented barriers such as high capital expenditure and production downtime. By design, our framework directly mitigates these primary obstacles, offering a pragmatic pathway for organizations to access these competitive advantages without undertaking disruptive and costly infrastructure overhauls.

This centralized approach also facilitates a more robust and seamless process for managing the model's lifecycle. Model retraining, updating, and versioning can be performed on the server infrastructure with zero disruption to the end-users. This enables a smooth transition between model versions, ensuring uninterrupted service and eliminating the version control and compatibility issues that are common in decentralized, on-device deployments. 

Furthermore, the framework's design provides a strong theoretical basis for predicting high user acceptance. Its reliance on a lightweight, browser-based interface aligns with the HCI principle of minimizing cognitive load by leveraging a universally familiar paradigm \citep{Hinze-Hoare2007-cm}. More critically, the design directly addresses the ``black box" problem, a major barrier to trust in AI, by affording experts direct control over model parameters. This interactivity embodies the core HCI principles of user control, feedback, and visibility, fostering a deeper understanding of the model's behavior and reinforcing the expert's role as the ultimate decision-maker. This design choice, coupled with the framework's positioning as an assistive tool that augments rather than replaces human expertise, is deliberately aimed at fostering the human-AI collaboration essential for adoption in a skilled workforce.

Beyond these benefits, hosting models centrally provides built-in advantages in scalability and security. The model-serving infrastructure can be scaled independently of the user-facing systems to meet environment-specific demand. Centralizing the models is also an effective security measure, as it prevents the expansion of proprietary model files across numerous devices, thereby safeguarding valuable intellectual property. Access can be tightly controlled and audited through the API gateway, ensuring that only authorized personnel can utilize the models.

Finally, the proposed framework is easily generalizable. While our initial prototype focuses on the NDI process within manufacturing, its principles can be readily applied to any scenario where specialized ML models need to be integrated into production environments with minimal disruption. This includes applications in fields as diverse as equipment manufacturing, food safety inspection, medical diagnostics, financial risk assessment, and legal document analysis. By providing a standardized interface for accessing diverse models, the API serves as a flexible foundation that can be adapted to the unique requirements of various industries, thereby promoting broader adoption and encouraging innovation.

\section{Limitations and Future Work}
While the proposed API-based framework presents a robust and pragmatic solution for integrating ML into legacy environments, it is important to acknowledge the boundaries of this work and identify areas for future investigation. The primary limitation is the absence of a formal, large-scale user study to empirically validate the usability and effectiveness of the prototype in a live production setting. Although the user interface was designed in accordance with established HCI principles to maximize user acceptance, future work should include quantitative studies to measure productivity gains, user satisfaction, and the impact on decision-making accuracy.

From a technical perspective, the framework's operational efficacy is contingent upon a stable and low-latency network connection between the on-site user interface and the remote API gateway. In environments with intermittent or low-bandwidth connectivity, the performance could be degraded, particularly when transferring large data files common in NDI tasks. Future iterations could explore edge computing solutions or intelligent data caching mechanisms to mitigate this dependency. Furthermore, while the API gateway centralizes security, it also introduces a new interface between the internal production environment and the remote model-hosting infrastructure, which requires careful security hardening and threat modeling beyond the scope of this initial paper.

Additionally, the current implementation serves as a functional prototype focused on the NDI use case. While the architectural principles are designed to be broadly generalizable, the specific user interface and data handling protocols have been tailored for this domain. Further research is needed to adapt and validate the framework's effectiveness across other industrial applications, such as predictive maintenance or supply chain optimization, which may have different data requirements and user workflows. These limitations provide a clear roadmap for subsequent research to build upon the foundational work presented here.

\section{Conclusion}
The challenge of integrating advanced ML capabilities into industrial sectors is not only a technical hurdle but a significant economic one. This often creates a barrier between cutting-edge innovation and established legacy systems. We proposed an API-based framework as a pragmatic and powerful solution to this problem. By decoupling the ML model lifecycle from the end-user environment, our approach eliminates the need for disruptive and costly hardware overhauls, simplifies model maintenance, and lowers the barrier to adoption for organizations of all sizes. This delivers the benefits of ML directly to the hands of domain experts through an accessible, browser-based interface. It mitigates financial and operational risks while fostering a more effective and trusting collaboration between human experts and intelligent systems.

The implications of the proposed method extend far beyond simple convenience, offering a strategic pathway for addressing pressing economic challenges, particularly in the context of domestic manufacturing. In regions where high production costs and a scarcity of specialized experts hinder industrial competitiveness, our proposed method serves as a powerful equalizer. It allows companies to leverage state-of-the-art AI without the prohibitive capital expenditure on in-house computational infrastructure. More importantly, it provides a way to significantly enhance the impact of human expertise. A central team of ML specialists can develop, deploy, and maintain models that serve numerous facilities, empowering on-site personnel to make better, data-driven decisions.

Furthermore, this approach is instrumental in enhancing production quality and operational safety. By embedding advanced analytical capabilities directly into the manufacturing workflow, organizations can implement rigorous, automated quality control and predictive safety monitoring. This ensures a higher, more consistent standard of output and reduces the likelihood of human error, which is critical in high-stakes industries. The resulting decrease in defects, material waste, and rework translates directly into lower operational costs and improved economic viability, making industries more competitive and sustainable.

The proposed API-based framework is more than just a technical framework. It enables a more agile, intelligent, and economically resilient industrial landscape. By democratizing access to Machine Learning, this method provides a clear and achievable roadmap for industries to enhance quality, ensure safety, and reduce costs, ultimately fostering innovation and strengthening their competitive edge.

\bibliography{references}

\end{document}